\documentclass[manuscript]{acmart}

\usepackage{enumitem}
\AtBeginDocument{%
  }

\setcopyright{acmcopyright}
\copyrightyear{2024}
\acmYear{2024}
\setcopyright{acmlicensed}\acmConference[To appear in  IUI '24]{29th International Conference on Intelligent User Interfaces}{March 18--21, 2024}{Greenville, SC, USA}
\acmBooktitle{29th International Conference on Intelligent User Interfaces (IUI '24), March 18--21, 2024, Greenville, SC, USA}
\acmDOI{10.1145/3640543.3645165}
\acmISBN{979-8-4007-0508-3/24/03}

\setcopyright{none}
\renewcommand\footnotetextcopyrightpermission[1]{}
\begin{document}

\title[Creating an African American-Sounding TTS]{Creating an African American-Sounding TTS: Guidelines, Technical Challenges, and Surprising Evaluations}
\author{Claudio Pinhanez}
\affiliation{%
  \institution{IBM Research}
  \city{São Paulo}
  \country{Brazil}
}

\author{Raul Fernandez}
\affiliation{%
  \institution{IBM Research}
  \city{Yorktown Heights, NY}
  \country{USA}
}

\author{Marcelo Grave}
\affiliation{%
  \institution{IBM Research}
    \city{São Paulo}
  \country{Brazil}
}

\author{Julio Nogima}
\affiliation{%
  \institution{IBM Research}
    \city{São Paulo}
  \country{Brazil}
}

\author{Ron Hoory}
\affiliation{%
  \institution{IBM Research}
  \city{Haifa}
  \country{Israel}
}


\begin{abstract}
Representations of AI agents in user interfaces and robotics are predominantly White, not only in terms of facial and skin features, but also in the synthetic voices they use. In this paper we explore some unexpected challenges in the representation of race we found in the process of developing an U.S. English Text-to-Speech (TTS) system aimed to sound like an educated, professional, regional accent-free African American woman. The paper starts by presenting the results of focus groups with African American IT professionals where guidelines and challenges for the creation of a representative and appropriate TTS system were discussed and gathered, followed by a discussion about some of the technical difficulties faced by the TTS system developers. We then describe two studies with U.S. English speakers where the participants were not able to attribute the correct race to the African American TTS voice while overwhelmingly correctly recognizing the race of a White TTS system of similar quality. A focus group with African American IT workers not only confirmed the representativeness of the African American voice we built, but also suggested that the surprising recognition results may have been caused by the inability or the latent prejudice from non-African Americans to associate educated, non-vernacular, professionally-sounding voices to African American people.


\end{abstract}

\maketitle

\section{Introduction}

Most of American English \emph{Text-to-Speech (TTS)} professional systems in use today produce voices which are distinctively recognized as White. This poses challenges for applications interested in targeting specific demographics (e.g., an African American business or NGO; a voice-tutoring system for children that are not of White ethnicity, etc.). The ultimate goal of the project described in this paper is to  provide to designers, developers, and enterprises the choice of having a professional voice which is clearly recognizable as African American, and therefore more able to address diversity and inclusiveness issues.


Being more precise, our goal is to create an \emph{African American Text-to-Speech system}, which we will refer simply as an \emph{African American voice  or AA voice}, able to produce synthetic audio segments from standard English texts, and which will be recognized by African American speakers and non-speakers as sounding like a native African American speaker. The AA voice should exhibit a level of technical quality similar to the \emph{Standard American English (SAE)} synthetic voices currently available through professional platforms. The evaluation of the technical quality of the AA voice, however, is not addressed in this paper, which focuses primarily on whether the AA voice can be recognized as sounding like an African American speaker. 

Linguists~\cite{McWhorter1998,McWhorter2017} have described a continuum of dialects under what is often termed \emph{African American Vernacular English (AAVE)}. At one end of the spectrum, one finds the largest deviation from SAE in terms of lexicon (including slang), syntax and morphology, 
and phonological/phonetic properties. At the other end, AAVE speakers begin to approach SAE in terms of lexicon and grammar but still retain marked speech characteristics (primarily in terms of intonation, phonation, and vowel placement~\cite{Green2002,McWhorter2017}) which grant the speech a distinctive identity which listeners use as cues in the perception of African American English~\cite{Walton1994}. 

There is some evidence that the race/ethnicity of a speaker of African American English can be determined solely by their voice~\cite{purnell1999perceptual,lattner2003talker,munson2007acoustic,chasaide2015speech}. \citet{purnell1999perceptual} performed a study where 20 speakers of three groups, African American, Hispanic American, and European American, recorded samples of the utterance \emph{``Hello, I’m calling to see about the apartment you have advertised in the paper.''} using \emph{African American Vernacular English (AAVE)}, \emph{Chicano English (ChE)}, and \emph{Standard American English (SAE)} respectively. Those samples were evaluated by 421~diverse participants, who correctly identified the race/ethnicity of the speaker more than 75\% of the time, ranging from 77\% to 97\% according to the speaker. In an identical experiment that just used the word \emph{``Hello''}, participants were correct about 70\% the time. A similar neuro-magnetic study with brain imagery~\cite{scharinger2011you} identified different areas of the brain being activated based on the dialect of the speaker.



At the heart of our work lies the need to collect the necessary resources to build a data-driven synthetic voice using the neural text-to-speech technologies we have previously developed~\cite{Shechtman-Fernandez:23}.
Since these systems are data driven, they are in principle capable of reproducing the subtle vocal characteristics of the data. Proper speaker selection is therefore crucial to successfully build the desired type of synthetic voice. 
To facilitate this, we carried a process of auditioning, narrowing down, and selecting a
professional voice which could satisfy our goals. 
That is, we sought to assemble a corpus of recorded English that grammatically
conforms to SAE but exhibits the phonetic and intonational features of AAVE to an extent that allows easy identification of the voice identity. 
We used a broad description of the intended target voice only to call for and audition voice actors for the project.
Final speaker selection, however, was empirically established by studies with the appropriate user demographics to explore the acceptability, likeability, and usability of the candidate voice.

This paper describes the process and 4~different user studies performed in the creation and evaluation of a synthetic voice for a professional \emph{Text-to-Speech (TTS)} platform that is expected to sound like an African American speaker. We start by presenting a study with 3~focus groups of African Americans which was conducted to elicit and discuss ethical issues and guidelines for the development of the voice. 
The same groups were used in a process to select the voice talent to be recorded to create the TTS system. 

We then describe how the voice was developed and some of the technical challenges faced by the team to capture the uniqueness of African American voices. Next, to evaluate the synthetic voice, we conducted three different studies, two with a demographically representative group of U.S. native English participants from a crowd-sourcing platform, and one with African American professionals. The studies showed, somewhat surprisingly, that the U.S. English speakers were unable to distinguish the African American synthetic voice as so. However, in the study with African Americans, there was clear certainty on attributing an African American ethnicity to the synthetic voice. When this latter group was shown the results obtained with the generic U.S. participants, they suggested that it was likely caused by misconceptions and prejudice of the participants, who might expect an African American speaker to make grammar ``errors,'' produce inadequate enunciation of words, speak \emph{Ebonics},\footnote{\url{https://www.linguisticsociety.org/content/what-ebonics-african-american-english}}, and use a limited vocabulary.







Considering the context and goals of the project, and the need to differentiate misconceptions of identity from non-African Americans, the research questions investigated in this paper are:
\begin{description}
    \item[RQ1:] Which are the key ethical considerations when developing an African American TTS synthetic voice?
    \item[RQ2:] Which criteria are important to select a typical and representative voice of an African American person?
    \item[RQ3:] Can current technology create a TTS generator which sounds like an African American person for African American speakers?
    \item[RQ4:] Can current technology create a TTS generator which sounds like an African American person for U.S. English speakers?
    \item[RQ5:] Can misconceptions and prejudices from the target audience affect the evaluation of the ethnicity or race of a synthetic voice?
\end{description}

The main contribution of this work is the documentation, and a discussion we will develop in detail as the paper unfolds, of this inherent tension between using inclusive human representations and avoiding negative stereotypes. Our results indicate that it is essential that the evaluation process of such systems consider different audiences not only because of distinct goals and socio-political agendas, but also because they may have different stereotypical representations of identities. 
Essentially, it is a paper that presents a case where that bias and prejudice impregnate all levels of technological development.


Before going forward, it is also important to point out the complexity of handling race-related terminology and some of the ways in which we address those issues. We acknowledge that the use of particular terms involves often debates and controversies across and even within communities. For instance, it is hard to provide a definition of what an African American person is, or how they sound, which simultaneously satisfies the many visions within the community while supporting in practical terms the discussions and studies described in this paper. We are aware of those issues and, as much as possible, we tried to be sensitive to the complexity of views and opinions inherent to them. 

Similarly, we are plainly aware that the availability of an African American TTS system opens up the doors for its inappropriate use, similar to female synthetic voices sometimes being used to utter sexist remarks. This is ethically deplorable but at the same time almost impossible to control. We see more harm being done, however, by simply continuing to not have the availability of high-quality, professional African American TTS systems.

\section{Related Work}

\subsection{Whiteness in AI and NLP}
The main context for this work is related to the debate about \emph{Whiteness of AI}. \citet{cave2020whiteness} discuss the prevalence of Whiteness in AI imagery and narratives, including robots, chatbots, and virtual assistants, drawing on \emph{Critical Race Theory}~\cite{delgado2011critical} concepts, particularly on \emph{White racial framing}~\cite{feagin2020white}. They also point out the consequences of this racialization of AI: amplification of prejudices; the portrayal of White machines as hierarchically superior to marginalized groups, and therefore justified in making decisions about them and thus 
affecting negatively disproportionally non-White people.

A lot of the work on racialization in AI has been performed in the context of robotics. The complicated relationship between humanoid robotics and race has its roots in the infamous \emph{Rastus Robot} created in the 1930s by Westinghouse, parading a Black blue-collar worker around the world to demonstrate its willingness to dangerous and humiliating stunts~\cite{abnet2020american}. Recent works have investigated the extent of Whiteness in humanoid robots~\cite{strait2018robots,sparrow2020robotics,bartneck2018robots,phillips2018human} and its implications~\cite{sparrow2020automation,cave2020whiteness}. Robots, even with limited humanoid features, were found to be seen as having race and, more often than not, portrayed as White. For instance, \citet{bartneck2018robots} found that machine-like robots with dark skin tones were as likely to be perceived as dangerous as Black human beings were in a shooter bias experiment. \citet{strait2018robots} showed that Black and Asian realistic humanoid robots attracted much more dehumanizing commentaries than a White equivalent.

There is also significant evidence that Whiteness is an important issue in text- and speech-based chatbots and virtual assistants: \citet{field2021survey} provide a good survey of different elements of race and racism in NLP and conclude, based on a study on the ACL anthology, that NLP has largely ignored race, and that the voices of historically marginalized people are nearly absent in the literature.

\subsection{Diversity in Synthesized Speech}

Research in speech synthesis has tackled several topics related to the development of synthetic voices in a non-predominant language variety. Examples of this include technologies for local minority languages that co-exist in a multi-lingual environment with a more predominant one, as is the case for regionally spoken languages in India~\cite{he2020open}; or endangered languages, like Irish Gaelic~\cite{chasaide2015speech}. Some work has looked at supporting code-switching during synthesis~\cite{thomas2018code}, which is one step in creating systems that resemble the vernacular of speakers from multi-lingual communities. All these examples deal with multilingual variety where the entities may be considered discretely distinct.

In the case of synthesis of dialectal variety within what may be broadly considered a single language, some work has looked at adapting from the standard form of the language to regional variants, usually at the prosodic level~\cite{pucher2010modeling,akiyama2018prosody}. African American English has been studied in linguistic departments~\cite{Green2002} but not received nearly as much attention in the speech-synthesis community, possibly due to the lack of available data resources needed to build systems.

In industry, there has been recent interest in bigger dialectal diversity, including African American English. The \emph{Acapela Group} announced in 2020 their intention to create \emph{``the first African American English Text-to-Speech Voices''}\footnote{\url{https://isaac-online.org/wp-content/uploads/Your-Voice-Matters-AssistiveWare-general-final-for-website.pdf}}, and their catalogue now 
includes 2 among 30 U.S. English voices which may be identified as such\footnote{https://www.acapela-group.com/voices/repertoire/}.
\emph{Apple} started to include, as of April 2021, two African American voices within their \emph{Siri} voice assistant, a fact that received a good amount of press coverage and online discussion\footnote{\url{https://www.consumerreports.org/digital-assistants/apples-new-siri-voices-resonate-with-some-black-iphone-users-a5978242346/}}.
There are other examples, narrower in scope, of creating dialectal variants, like \emph{Amazon}’s involvement in creating a Southern-U.S. synthetic voice for \emph{KFC} branding in Canada\footnote{\url{https://aws.amazon.com/blogs/machine-learning/build-a-unique-brand-voice-with-amazon-polly/}}.

\subsection{Impacts of Race and Cultural Background on User Conversational Experience}

Issues of racialization and cultural backgrounds in chatbots have been considered since the early days of AI. As pointed out by~\citet{marino2014racial}, the racialization of chatbots started with the very first chatbot, \emph{ELIZA} and its standard middle-class, heterosexual cues. Early work with \emph{Kyra} by~\citet{maldonado2004toward} already showed milder interactions for her Latina and Brazilian versions than for the American one. 

\citet{marino2014racial} discussed the interesting case of racialization of \emph{ANACAONA}, a chatbot designed to provide information about Caribbean Amerindians. The depiction of the chatbot as a modern-looking young woman attracted both abuse and criticism about her representation as a legitimate interlocutor of Amerindian culture. 
\citet{liao2020racial}, studying psycho-therapeutic text chatbots, found that the user’s perception of interpersonal closeness, satisfaction, and similar interaction attributes were higher when they had the same racial identity as the chatbot, whereas disclosure comfort was lower when chatbot and robot had different racial identities. 
In particular, the preference of African American participants for African American chatbots was the strongest among all racial groups.


In the specific context of speech, some studies have shown that listeners can rapidly categorize speakers in terms of dialect and race~\cite{purnell1999perceptual,lattner2003talker,munson2007acoustic,scharinger2011you}. \citet{arjmandi2018investigation} identified key distinguishing acoustic dimensions between Standard and African-American English, confirming observations in~\cite{arjmandi2017applying}. \citet{farrington2021sources} studied the sources of variation in the speech of African Americans. 
\citet{nass2005wired} explored various aspects of voice and text in machines and interfaces, including an experiment where participants demonstrated a decrease in trust when experiencing voice and text content which did not match the racial depictions of the chatbot (using ``typical'' White Australian and Korean ethnic faces). 

\citet{tamagawa2011effects} studied the effects of having healthcare robots with the same or different accents as their users and found that robots with similar accents were perceived as better. 
\citet{mayfield2019equity} described several cases, positive and negative, where issues related to race affected NLP applications, including a case where kids responded better to culturally-congruent voices in reading and comprehension tasks. 
\citet{cambre2019one} suggested a framework for understanding the social implications of voice design.

\section{Initial Study: Development Guidelines, Selection Criteria, and Voice Talent Ranking}\label{sec:ethical_issues}

The first study we describe here had as its main goal determining ethical concerns to be used in the development of the AA voice.  Since the TTS system would be based on the recordings of a single human voice reading a script, we also used this study to collect selection criteria for a representative voice, and to rank professional voice candidates previously auditioned in a recording studio in order to select the final speaker. Notice that the AA synthetic voice was expected to resemble, in many ways, the voice of the recorded person.

\subsection{Auditions}
\label{ssec:audition}


Open auditions were held in a recording studio in New York City. The call for talent included a statement describing the objective of developing 
\emph{``a high-quality, professional, artificial (synthetic) voice that is perceived as being that of a native speaker of African American English.''} The auditions were described as aiming to record \emph{``a few hours of audio that reflect general standard English usage 
 \ldots  from a speaker of African American English (as exemplified, for instance, by Black voices in media, news casting, etc.)''}. No other speaker characteristics were mentioned, nor gender or age. 

The studio recruited the talent and scheduled individual 15-minute sessions 
where we asked each talent to read a script containing two short dialogues, one brief narration, and eight isolated sentences. In total, 19 candidates took part in the audition (16 males and 3 females). 
The audition recordings were first examined by the team and narrowed down to a set of 11 speakers based on technical/practical issues.

\subsection{Focus Groups Methodology}

We run 3~focus groups lasting about 75~minutes each: the first two focus groups ranked only half of the candidates; then their respective top ranks were merged and re-ranked by the last focus group. Participants were African Americans recruited from an African American diversity \emph{Slack} channel of an IT company. They were told the basic premises and objectives of the project before agreeing to participate, and all consented to be recorded and to the use of the results of the study by the authors. They did not receive any reward for participating in the study.

Each focus group performed the following sequence of tasks: 
\begin{description}[labelindent=0cm,leftmargin=0.3cm,style=unboxed]
    \item[Introductions:] Participants were informed about the objectives of the study, the main steps, and asked to briefly introduce themselves.
    \item[Description of the project:] The organizers described the project, its goals, and the process of developing a synthetic voice.
    \item[\emph{Hopes and Fears} dynamics:] Participants were asked first to explore together the positive sides of the project and wanted outcomes (``hopes'') and then asked to think about what could go wrong with and undesirable outcomes (``fears'').
    \item[Listening and questions about the candidates:] Each participant 
    answered 6~questions, using a 5-level Likert scale, about different qualities of each candidate. 
    \item[Discussion about the task and selection criteria:] The participants were asked to comment about the difficulty of the task, and about which criteria they believed was appropriate to evaluate and rank the voices.
    \item[Group ranking of the candidates:] The participants were given an aggregate score for all candidates (based on the individual answers of each participant) and worked together to reach a consensus ranking of the candidates as the best choices for the AA voice. 
    \item[Thanks:] They were thanked and asked for suggestions and comments.
\end{description}

\subsection{Results and Findings}

The focus groups were performed in September of 2022 with 7, 4, and 6 participants in each group, all African Americans (9 female, 8 male). To refer to individual participants, we use the notation $Pi.j$ where $i$ is the number of the study and $j$ is an index randomly assigned to each participant.

The ``Hopes and Fears'' dynamics of the focus groups collected many ideas and ethical concerns about the project and its outcome. We processed the transcriptions, 
segmented the main ideas discussed, and grouped them into main themes. 
The main themes were:

\begin{description}[labelindent=0cm,leftmargin=0.3cm,style=unboxed]
\item[Ensuring participation of African Americans in all parts of the process.]
P2.2 expressed how important they felt it was for African Americans to be part of the process: \emph{``I like the inclusiveness of [people from our company] to be a part of the analysis, 
 \ldots  making sure that, when we're creating our models, that they are authentic. But using our people to be a part of that validation process is brilliant'' (P2.2)}. Similarly, P2.5 stated that \emph{``the main fear would be that you didn't do this focus group'' (P2.5)}.

\item[Addressing the difficulty of representing diversity of African American voices.] An important concern was how a single AA voice would be able to represent the diversity of African American voices, and \emph{``[how] to distill a voice that reflects the African American community and the different dialects and different sounds that we have''~(P1.7)}. P2.1 expressed some doubt about the feasibility of the task, \emph{``the breadth of African American experience is so large \ldots . What do you want that voice to represent for this community?'' (P2.2)}, while P3.3 was concerned that \emph{``depending on what region you're in, for African Americans, you can't tell just by talking with them if they're African American or not'' (P3.3)}.

\item[Avoiding the different forms of stereotyping and the use of Ebonics.] Concerns about the AA voice following exaggerated or negative stereotypes of how African Americans speak were very strong. P3.2 made that clear, \emph{``I am concerned that they're going to use a stereotypical voice as well. So give us representation, but don't give us too much representation'' (P3.2)}. 
Similarly, there were important concerns about the AA voice speaking Ebonics, \emph{``I~just don't want to see this project go down that path [of Ebonics]'' (P2.3)}. And as P2.4 put, 
\emph{``the main point is it must be respectful and authentic, if it wasn't respectful or authentic  \ldots  that could lead to Ebonics'' (P2.4).}

\item[Handling the difficulty of non-African Americans to distinguish African American voices.] Participants pointed out the difficulty that non-African Americans have in recognizing, in audio-only contexts, that a person is African American. P3.6 told  
that \emph{on the phone  \ldots  people think that I come across [as White], especially with my name'' (P3.6)}. P3.1 and P3.2 produced similar accounts. This issue also happens with another ethnic and racial groups: \emph{``When I volunteer over the phone with a minority [non-African American] community, they can't see my face. There's little connection because they think I am White. 
 \ldots  it's totally different, though, when they can see me''~(P3.6)}. 
They also joked about their ability to control the recognition: \emph{``When I put my James Earl Jones\footnote{An African American actor known for his low pitch, deep voice.} voice, meaning more more deeper-tone qualities, then I come off as being African American''~(P3.1)}.

\item[Assuring that the AA voice is used and measured correctly.] There were also concerns about \emph{``how [the AA voice] will be perceived and used. I don't think it means don't do it, but I think we have to realize what the pros and the cons could be'' (P3.5).} Similarly, what could happen if the AA voice was not a success, \emph{``will this be, like, part of a metric in the sense of how many people are using this voice?'' (P1.5).} 
\end{description}

In the discussion about the criteria for used for ranking, conducted between the individual and group ranking phases, 4 main criteria were mentioned:
\begin{description}[labelindent=0cm,leftmargin=0.3cm,style=unboxed]
    \item[Representativeness.] As P2.3 put it, \emph{``The goal I thought for this was to have the synthetic voice represent African Americans'' (P2.3)}. Regarding the final group of candidates, \emph{``I thought there was safe candidates, I think they were decent representations'' (P3.1)}.
    \item[Clarity and speech quality.] The need of good enunciation was mentioned often, \emph{``I think I focused more on the pronunciations, I want to make sure that no matter what voice was chosen it was able to articulate and pronounce every syllable of every word, as well as the cadence and the delivery'' (P1.5)}. Similarly, P2.1 stated that \emph{``The 3 things that I was looking for were clarity, articulation and pacing'' (P2.1)}.
    \item[Empathy and expressiveness.] P1.7 pointed out that they \emph{``tried to look at the empathy of it'' (P1.7)}, while P2.2 said that the \emph{``choice was made on how I was made to feel when I heard the voice. That I have a feeling of comfort, of joy, of familiarity, of connectivity, of sameness'' (P2.2)}.
    \item[Appropriateness to its use.] Another criteria was whether the talent voice would be appropriate to the two scenarios of the study, customer service and tutoring: \emph{``
    [Some voices] didn't really have a strong command presence for customer service or a request, as opposed to reading someone a book. So I tried to think of the voices that were the most utilitarian for every aspect as the criteria'' (P1.7)}. Similarly, P2.1 commented that \emph{``when it came to customer service, I was looking for an expression of empathy, and then when it came to reading, could that person hold my attention?'' (P2.1)}.
\end{description}

In the listening and evaluation phase, all three groups produced lists with clear distinctions between the best and worst candidates. The participants selected 4~good candidates in group~1, 3~in group~2, and 4~candidates were on the top in the final round. After some debate, the participants agreed that the top ranked candidate, a female talent, was the best choice. This voice talent was the one who recorded the training data for the AA voice and some audio samples of her voice can be heard in the supplementary material\footnote{\url{https://github.com/cpinhanez/iui24aatts}}.

\subsection{Discussion}

The 3~focus groups not only provided the project with a strong, representative voice talent to record the training data, but also much-needed development guidelines and evaluation criteria.  

Regarding \textbf{RQ1}, the focus groups offered five key guidelines to be followed by the project, as discussed above:
\begin{enumerate}
    \item \textbf{Ensuring participation of African Americans in all parts of the process.}
    \item \textbf{Addressing the difficulty of representing diversity of African American voices.}
    \item \textbf{Avoiding the different forms of stereotyping and the use of Ebonics.}
    \item \textbf{Handling the difficulty for non-African Americans to distinguish African American voices.}
    \item \textbf{Assuring that the AA voice is used and measured correctly.}
\end{enumerate}

The participants also provided the project with a set of criteria for an African American voice, answering \textbf{RQ2}:
\begin{enumerate}
    \item \textbf{Representativenes.}
    \item \textbf{Clarity and speech quality.}
    \item \textbf{Empathy and expressiveness.}
    \item \textbf{Appropriateness to its use.}
\end{enumerate}


\section{Creating a State-of-the-Art TTS System from Scripted Recordings}

We describe next the main steps followed to record and develop an AA voice from the voice talent chosen by the African American focus group participants as previously described.

\subsection{Creation of a TTS Conversational Corpus: Recording Script and Recordings}

To enable the development of the AA voice, we created an in-house corpus of expressive conversations from studio recordings of the African American female speaker of U.S. English selected in the studies described before.  
This corpus primarily  consists of dialogues between a customer-care AGENT and a USER, containing various emphatic items, as well as various interjections (e.g., {\em ``oh''}, {\em ``aha''}) and filled pauses (e.g., {\em ``um''}, {\em ``uh''}) that are characteristic of spontaneous conversation. 

Two opposing goals are at play when developing a corpus of conversations for TTS: on one hand, we value a style that feels spontaneous and not overly acted. Purely spontaneous or ad-libbed speech, however, can suffer from a variety of artifacts (e.g., ill-formed phrasing, stuttering, repetitions, etc.) which are difficult to model well with existing technology without compromising its quality. Some proposals to address this during data collection include two-pass approaches where the talent is given a chance to rehearse, followed by a final improvisation pass~\cite{Guo-Zhang:21}. 

As this can be time-consuming and expensive, we instead took a {\em role-playing approach} by crafting dialogues {\em a priori} that are carefully designed to contain and elicit expressive situations, and are also carefully annotated with dialog-act tags that are available to the actor to exploit while performing her reading. When recording, the talent has access to the entire history of the dialogue as it gradually takes shape, including both the AGENT turns (which they recorded) and the responses from a USER (which they silently read and are given a chance to react to on their next turn). Giving the actor access to turns in context can elicit a wider range of expressiveness and naturalness than recording isolated turns. The talent was  also allowed the freedom to make minor deviations from the script, and the script was corrected {\em a posteriori} to conform to the audio. 

For the AA voice, the actor was first made aware about the intended technological use of her recordings. Recordings sessions took place in the same studio as the auditions, totaling 16 hours of studio time. This amounts to about 6 hours of actual speech, of which 5.5 hours are reserved for training and the rest  held out as validation and test.  An instructional session with the actor was held at the beginning to go over and rehearse the intended target conversational style, and the meaning of the various annotations in the script. To avoid biasing the actor's performance toward a more {\em standard} performance, she was given only broad stylistic guidelines about the meaning of the expressive labels. 
The same was done for phonetic content: issues of word pronunciations were only addressed in a few cases in which the actor consulted with the producing team about the pronunciation of a rare word or proper name, and in such cases a canonical dictionary pronunciation was suggested to her.

\subsection{Overview of the TTS System}

\begin{figure}[t!]
\vspace{-2mm}
\centering
\centerline{\includegraphics[width=\columnwidth]{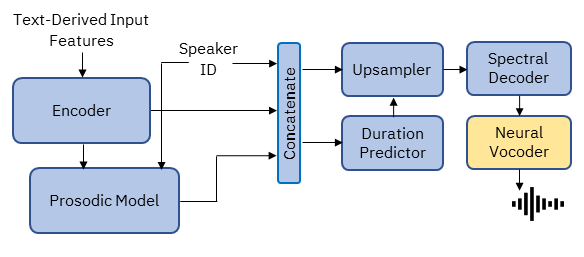}}
\setlength{\abovecaptionskip}{8pt}
\setlength{\belowcaptionskip}{0pt}
\caption{Text-to-Speech Architecture. An acoustic model (blue) predicts spectral sequences from input text, which are then sent to an independent neural vocoder (yellow) to generate final speech.}
\vspace{-2mm}
\label{fig:ttsarch}
\end{figure}

The TTS architecture used in this work (see fig.~\ref{fig:ttsarch}) consists of two components: (i) an {\em acoustic} model that transforms input text, augmented with annotations, into a set of acoustic spectral features, and (ii) a vocoder that converts these predicted features into an audio waveform. The acoustic model in our case is a sequence-to-sequence, encoder-decoder, neural architecture largely based on the \emph{Non-Attentive Tacotron2} models proposed in~\cite{Shen-Pang:18,Shen-Jia:20} to predict a set of spectral features from text, but modified to also contain a prosodic model that predicts in tandem a set of prosodic-acoustic measures (similar to those described in~\cite{Shechtman-Fernandez:23}) 
which play a crucial role in accurately modeling the range of expressiveness of conversational data.

The encoder component consumes, and processes for both the decoder and the prosodic module, various types of symbolic features including (i) linguistic information provided by an independent text-processing front-end (e.g., phonetic sequences, types of phrases);  and (ii) various metadata characterizing the content of utterances (e.g., discourse-level dialog-act annotations; the presence and type of interjections and filled-pauses; etc.). The encoder's output is merged with the speaker's ID (to support multi-speaker models) and fed into the prosodic module to predict a set of pitch, duration, and energy measures evaluated at various hierarchical prosodic spans (e.g., words and utterances). All this information is then merged and passed to (i) a duration model that determines how long each input phonetic unit should be, and (ii) a decoder that predicts the output spectral sequences. To obtain the final speech waveforms, the spectral sequences predicted by the acoustic model are then processed by the vocoder, an independently trained speaker-specific neural \emph{LPCNet} model~\cite{Valin-Skoglund:19}.

\subsection{Generating the Synthetic Prototypes}

AA voice models were developed using a mature training recipe previously validated using various other (non-African American) voices in U.S. English, as well as several other languages. Both \emph{single-speaker (SS)} and \emph{multi-speaker (MS)} models were considered when training the acoustic model component, although the vocoder component was, in all cases, a speaker-dependent model trained exclusively on the voice of the African American talent. A MS approach uses data from multiple speakers, multiple styles, and, in this case, multiple dialects; this framework has been shown to lead to models with better and more stable quality (even when the intended run-time target speaker and target styles differ from those of the extra data used at training time) while also preserving the identity of the run-time speaker, since speaker information is explicitly modeled in the architecture as a conditioning variable, as fig.~\ref{fig:ttsarch} shows. 

Since  preservation of the dialectal features of the target speaker is fundamental to this work, we found it worthwhile to compare SS models (trained only on 5.5 hrs of audio)  to MS models trained with an additional 60 hours of audio from 4 other non-African American speakers of SAE (3 females and 1 male).  
The findings of this procedure based on informal listening by speech experts (using an independent set of sentences not used during training) was that the MS approach indeed compensated for some articulation and prosody artifacts that had been observed in the SS models developed with fewer resources\footnote{This comparison was not done via formal listening tests.}, but we include samples from both in the supplementary material to illustrate the improvements\footnote{\url{https://github.com/cpinhanez/iui24aatts.}}.
For this reason, the MS model with the best validation loss during training was selected to generate the candidate sentences for the recognition studies described next. 

Using this model, 10~text sentences which syntactically and lexically conform to SAE and not seen during training were used to synthesize the 10~speech segments employed in the studies. These customer-care domain texts reflect the AGENT response in representative interactions and include dialogue devices like filled pauses and interjections.  


\begin{figure*}[t!]
  \includegraphics[width=13cm]{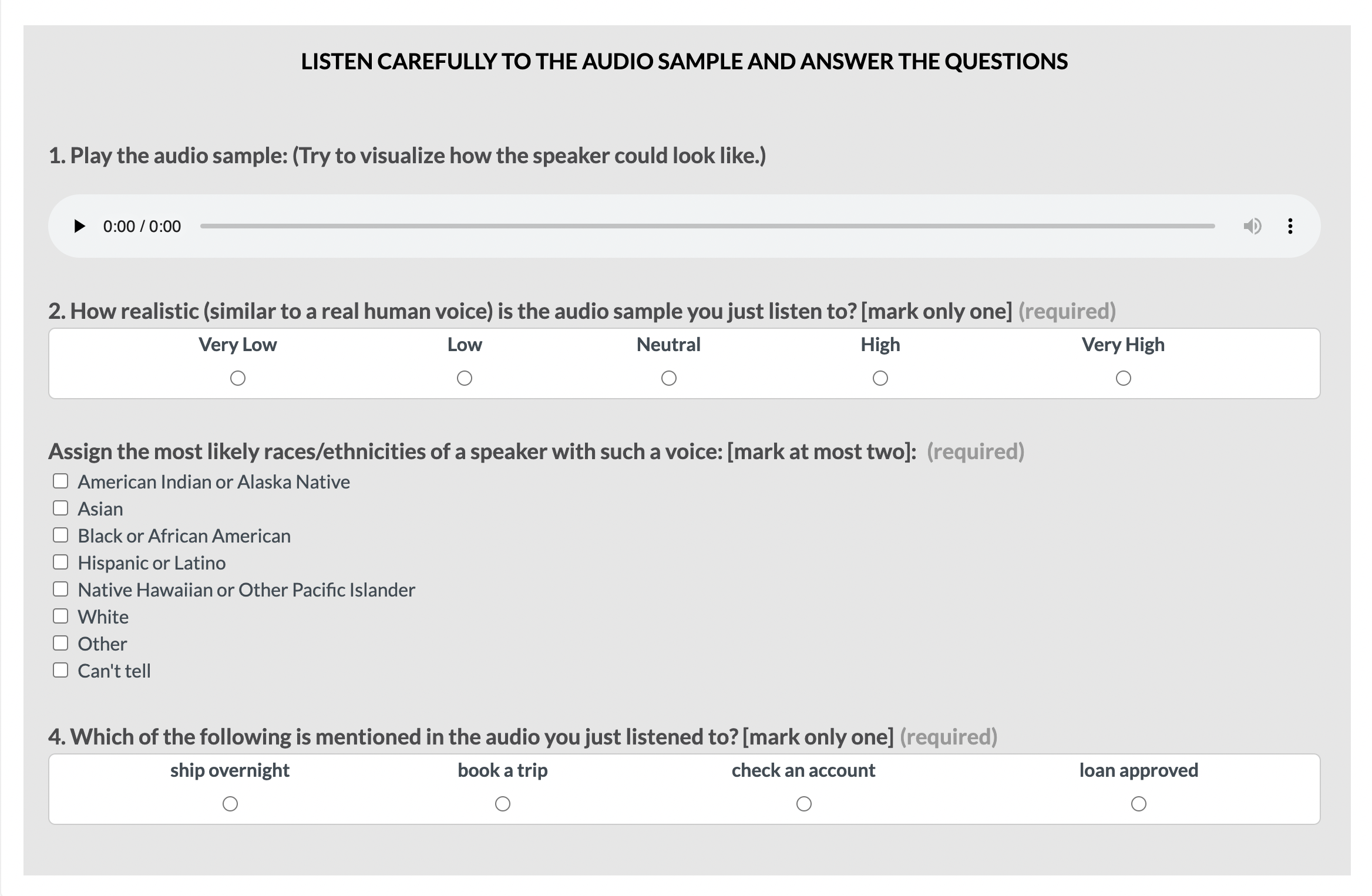}
   \caption{Screen interface of the main task performed by the participants of the Study~1.}
   \label{fig:study1_screen} 
\end{figure*}

\section{Evaluating Whether the AA Voice Sounds African American}

To evaluate whether U.S. English speakers recognized the created AA voice as being from an African American person, we performed two different studies with about 570 participants who performed more than 2,848 evaluations. Participants in these studies were digital workers from the crowd-sourcing platform \emph{Appen}\footnote{https://appen.com}. We set up the experiment as a task, paying approximately the minimum wage of California, and only digital workers from the U.S were allowed to participate. The validity of the use of digital workers as participants in HCI studies similar to ours has been established by several studies~\cite{kittur_et_al_crowdsourcing_studies_2008,liu_et_al_crowdsourcing_usability_testing_2012,daniel_et_al_crowdsourcing_quality_2018,hauser_etal_MTurk_2018,jimenez_et_al_reliability_crowdsourcing_2018,jimenez_et_al_stimuli_speech_quality_2018}.

In the studies we also used, as a comparative baseline, another synthetic TTS voice of comparable quality, created by us with the same technology based on the recordings of a White woman. We refer to that voice simply as the \emph{White voice (WH voice)}. Using 10~text sentences from the customer service domain which syntactically and lexically conform to SAE,  we synthesized 10~speech segments for each voice to be used in the studies, available for listening in the supplementary material\footnote{\url{https://github.com/cpinhanez/iui24aatts}}.
In the studies the participants never directly compared the voices to each other.

\subsection{Study~1: Direct Question}

The goal of \emph{Study~1} is to determine whether U.S. English speakers attribute the AA voice, if it was a person, to an African American, by asking a direct question about it. This question was constructed using the race and ethnicity categories defined by the \emph{Standards for the
Classification of Federal Data on Race
and Ethnicity} of the \emph{U.S. Office of Management and Budget (OMB)} in 1997, and used in the U.S. Census since 2000\footnote{https://www.govinfo.gov/content/pkg/FR-1997-10-30/pdf/97-28653.pdf}. According to this standard, at least 6 categories for race and ethnicity should be used: \emph{American Indian or Alaska Native}, \emph{Asian}, \emph{Black or African American}, \emph{Hispanic or Latino}, \emph{Native Hawaiian or Other Pacific Islander}, and \emph{White}.

\subsubsection{Study~1: Methodology} 
\hfill \\
Participants were told that they were evaluating the quality of a synthetic voice and, in the process, they were also asked about the race and ethnicity of the voice. We decided to disguise the study to minimize the effect of personal opinions about race and ethnicity. 
Participants were equally exposed to both the AA and WH voices uttering 5 to 10-word sentences typical of dialogues in customer service. The demographics of customer-care centers basically follows the overall U.S. demographics\footnote{https://www.zippia.com/customer-service-representative-jobs/demographics/},  so we believe the use of customer service content did not affect the findings of our studies.

Each participant in Study~1 performed the following sequence of digital tasks:
\begin{description}[labelindent=0cm,leftmargin=0.3cm,style=unboxed]
    \item[Introduction:] Participants were informed about the objectives of the study, the main steps, provided with an example of the main task, and given rules and tips for success.
    \item[Main task:] Participants performed 4 times a process of listening to an audio sample, evaluating its realism, choosing the most likely race/ethnicity of the speaker, and answering a golden question about the content 
    (see fig.~\ref{fig:study1_screen}).
    \item[Thanks:] Participants were thanked and asked for suggestions and comments.
\end{description}

\begin{table}[t!]
  \caption{Total numbers of digital workers and evaluations in both studies and their separation into valid and invalid.}
  \label{tab:study_numbers}
  \includegraphics[width=\columnwidth]{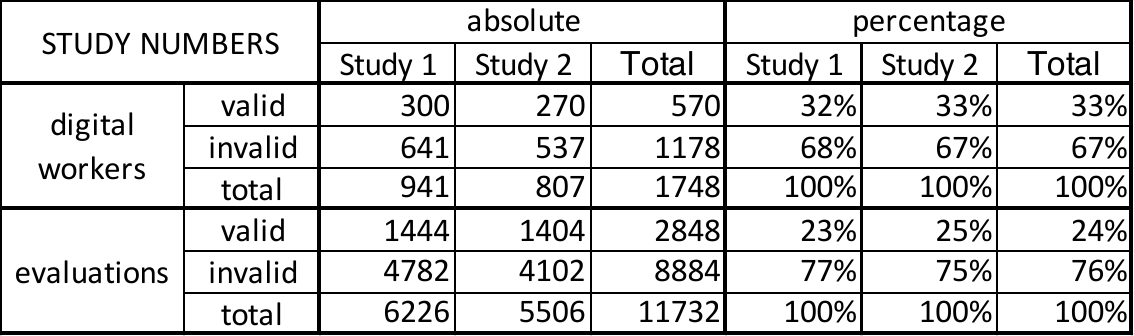}  
\end{table}

\begin{table*}[t!]
  \caption{Results of Study~1 where participants were asked to attribute the AA voice audio samples $AA_i$ and the WH voice audio samples $WH_i$ to the ethnicity/race categories shown on the first column of the table; the most selected races and the overall most chosen race are marked with a grey background.}
  \label{tab:results_study1}
  \includegraphics[width=\textwidth]{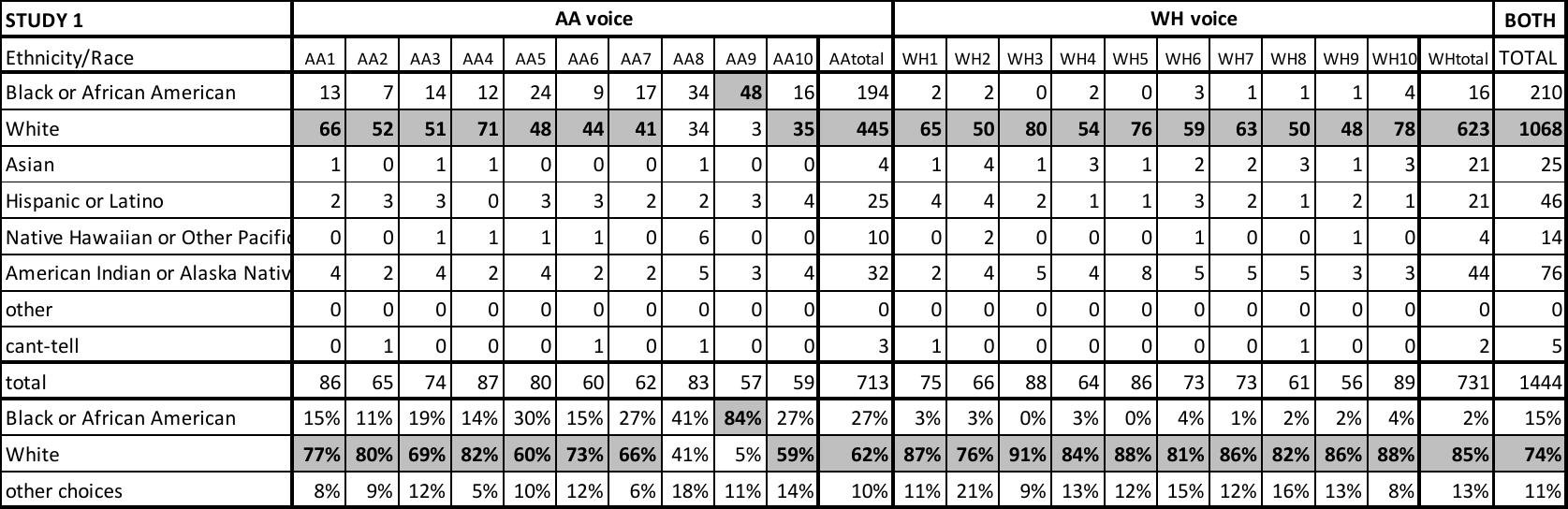}
  
\end{table*}

Figure~\ref{fig:study1_screen} shows the screen interface of the main task performed by the participants of Study~1. Notice that the participant has to first answer the disguise question about realism, then follow with the multiple-choice question about race and ethnicity, and end with the golden question about the content of the audio sample. Answers about race and ethnicity from participants who failed the golden question were automatically discarded since it could be evidence that they had not attentively listened to the audio.
Participants were randomly assigned to different audio samples spoken by the AA voice and by the WH voice, 2 from each voice. Randomness was controlled so they would never listen to the same text spoken by both voices.

\subsubsection{Study~1: Results and Findings} 
\hfill \\
As detailed in table~\ref{tab:study_numbers},
a total of 941~participants performed Study~1, of which 300 (32\%) were considered valid. Of the 641 who were considered invalid, 15 failed the quiz and 626 were deemed as bots due to multiple, identical evaluations occurring at the same time. The participants 
performed 6226~evaluations in Study~1, of which 1444 were considered valid (23\%). There were 713~evaluations for audio samples of the AA voice and 731~evaluations for the WH voice. 
Since this study was performed within 3 weeks of the process of selecting stereotypical faces described in appendix~B, we believe that it likely follows the same demographics, which we found in that study to closely resemble the demographics of the U.S.

Table~\ref{tab:results_study1} displays the overall results of Study~1. It shows the race/ethnicity attributions to each of the 10~audio samples of the AA and WH voices. 
The percentages shown in the three bottom rows correspond to the total number of evaluations, where \emph{other} corresponds to all options which were not either ``Black or African American'' or ``White''. 
We marked in bold and with a grey background the race/ethnicity selected by most of the participants. 
For the AA voice audio samples, 7~were attributed to a ``White'' person and only one to a ``Black or African American''. There was a predominant attribution to ``White persons'', with 62\% of the evaluations against 27\% to a ``Black or African American'', and 10\% to other ethnicities/races. 

Notice that in the case of the WH voice, all audio samples were attributed to a person of a White race in more than 85\% of the evaluations 
and only about 2\% of the evaluations attributed the WH voice samples to an African American person while 13\% were attributed to other ethnicities/races.
Those were quite unexpected results, but we defer the discussion about possible reasons for this and other findings until after we have presented Study~2 and Study~3, and their respective results.

\subsection{Study~2: Face Matching}

In parallel with Study~1 we performed a second study, referred to as \emph{Study~2}, to evaluate the AA voice using a different methodology but within the same basic population of participants. 
The evaluation task of Study~2 was matching audio samples of the voice to stereotypical faces of different races.

\subsubsection{Study~2: Methodology}
\hfill \\
In Study~2 participants were asked to match the same audio samples used in Study~1 to one of 6~stereotypical, synthetic female faces of three different races: African American (noted as \emph{AA}), White (\emph{WH}), and Asian (\emph{AS}). The issue of race or ethnicity was never mentioned in the tasks performed by the participants. 
Study~2 was inspired by the work of~\citet{nass2005wired} which showed that a mismatch between the race or ethnicity of a voice and a portrait of the speaker has impacts on users' trust. 

To perform Study~2, we first had to create a set of stereotypical AA, WH, and AS female faces. This process, detailed in appendix~A,
started with the selection of the \emph{Synthetic Faces High Quality (SFHQ)} dataset\footnote{https://github.com/SelfishGene/SFHQ-dataset} as the source of images. This dataset comprises more than 425,000 synthetic faces, classified by race and ethnicity and several other characteristics. We first extracted about 200 faces of each race, and then each of the authors selected 20 faces from them considering criteria such as image quality, age between 25 and 40 (considering the characteristics of the voices), stereotypicity, and whether the face was smiling. For each race the set was consolidated by removing any disagreements; this resulted in 20 faces that were then submitted to a crowd-sourcing process where faces were approved only if at least 9 in 10 participants assigned that face only to a single race. This led to the set of 10 faces of each race depicted in fig.~\ref{fig:faces_study2}. More details, including ethical considerations, are given in appendix~A.

\begin{figure*}[t!]
  \includegraphics[width=15cm]{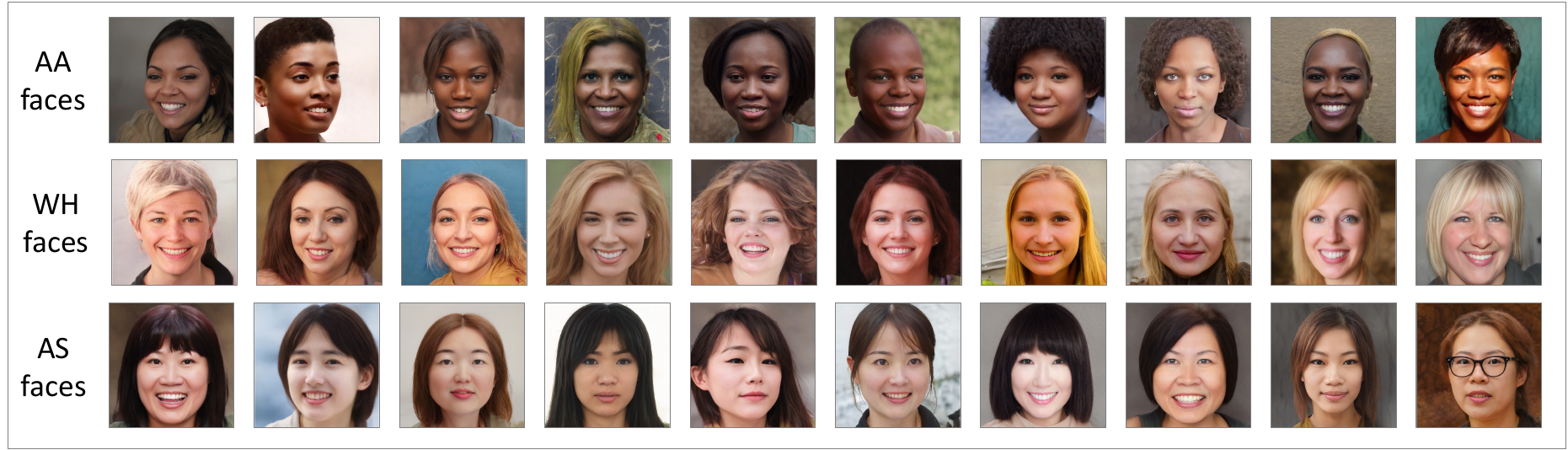}
   \caption{Synthetic faces used in Study~2.}
   \label{fig:faces_study2} 
\end{figure*}

\begin{figure*}[t!]
  \includegraphics[width=15cm]{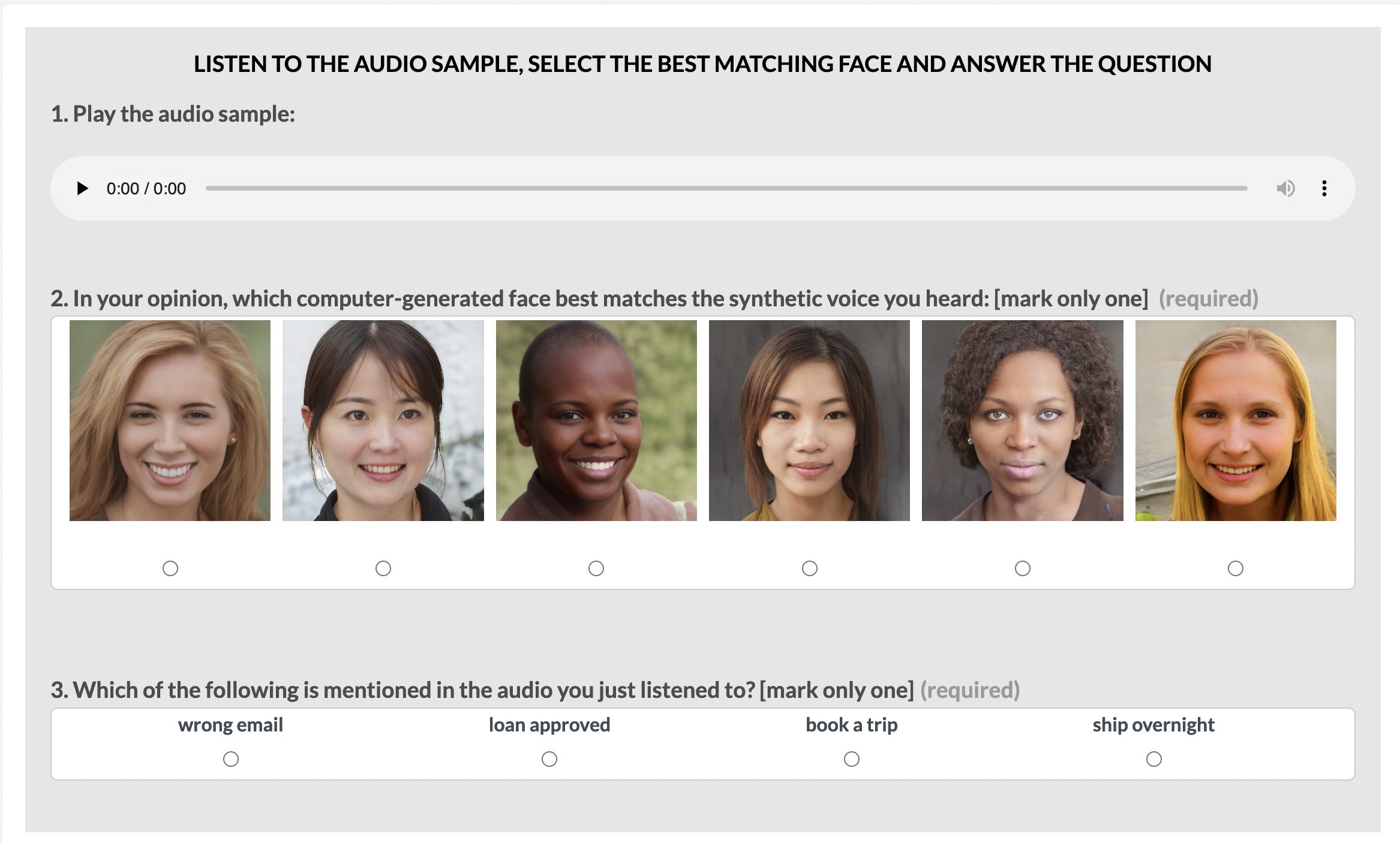}
   \caption{Screen interface of the main task performed by the participants of the Study~2.}
   \label{fig:study2_screen} 
\end{figure*}

Participants in Study~2 were also digital workers from the crowd-sourcing platform \emph{Appen} under the exact same conditions and constraints of Study~1. Participants were told that they were helping the selection of a synthetic face as an avatar for a synthetic conversational agent. The audio samples were exactly the same ones used in Study~1.

Each participant in Study~2 performed the following sequence of tasks:
\begin{description}[labelindent=0cm,leftmargin=0.3cm,style=unboxed]
    \item[Introduction:] Participants were informed about the objectives of the study, the main steps, provided with an example of the main task, and given rules and tips for success.
    \item[Main task:] Participants performed 4 times a process of listening to an audio sample, selecting the best match among 6 synthetic faces, describing difficulties (every two tasks), and answering a golden question about the content 
    (see fig.~\ref{fig:study2_screen}).
    \item[Thanks:] Participants were thanked and asked for suggestions and comments.
\end{description}

Figure~\ref{fig:study2_screen} shows the screen interface of the main task performed by the participants of Study~2. Notice that the participant has to choose the face which they think best matches the synthetic voice in the audio sample. 
Answers  from participants who failed the golden question were automatically discarded like in Study~1.
Participants were randomly assigned to different audio samples spoken in the AA and WH voices, in groups of 4 (2 from each voice). Randomness was controlled so they would never listen to the same text spoken by both voices. Similarly, the faces were randomized so a participant would never see the same face twice. In each matching task, there were always two faces from each race.

\subsubsection{Study~2: Results and Findings}
\hfill \\
As detailed in table~\ref{tab:study_numbers},
a total of 270~participants (excluding 547 which were considered invalid) performed 5506~evaluations in Study~2, of which 1404 (25\%) were considered valid. There were 704~evaluations for audio samples of the AA voice and 700~evaluations for the WH voice. This study was performed at the same time as Study~1, so we believe the demographics of the participants are essentially the same.

\begin{table*}[ph!]
  \centering
    \caption{Results of Study~2 where participants were asked to match the AA voice audio samples $AAi$ and the WH voice audio samples $WHi$ to sets of stereotypical AA, WH, and AS faces; the most selected faces and the overall most chosen race are marked with a grey background.}
  \label{tab:results_study2}
  \includegraphics[width=0.98\textwidth]{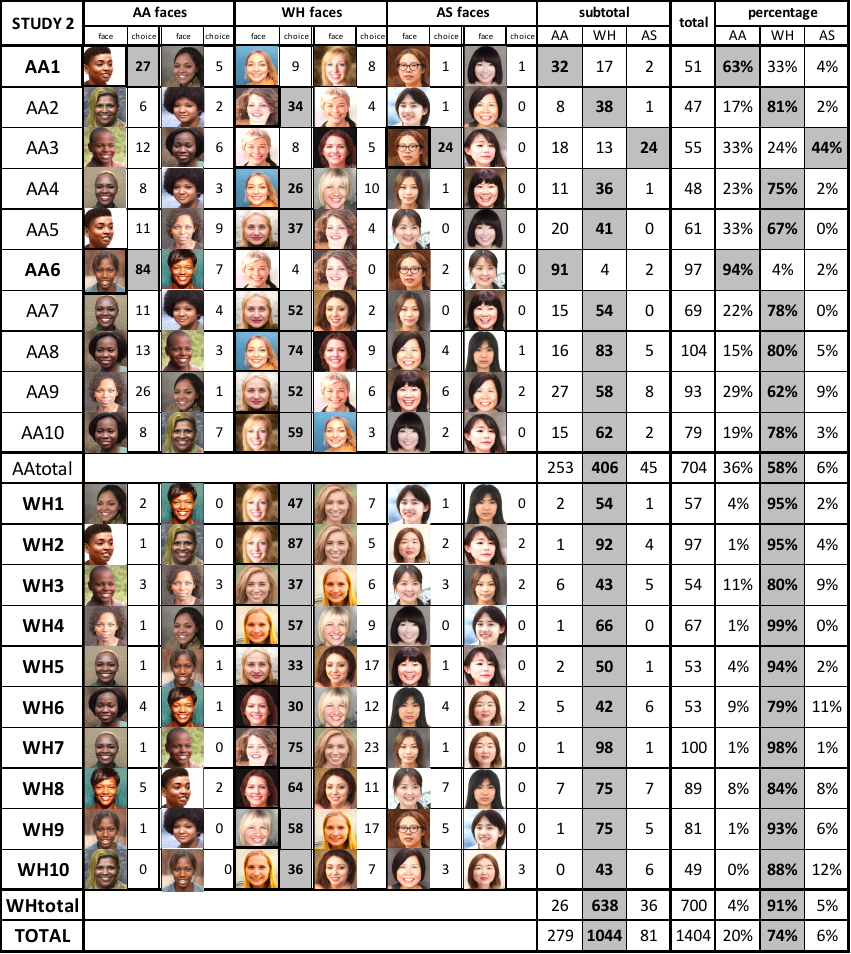}

\end{table*}

Table~\ref{tab:results_study2} displays the overall results of Study~2. It shows the number of times each of the 10~audio samples of the AA voice and of the WH voice was matched to each face of the corresponding set of AA, WH, and AS faces. 
Notice that no audio sample was intended to match an AS face, although in about 6\% of the evaluations they were ``incorrectly'' matched by the participants.

In table~\ref{tab:results_study2} we marked in boldface and with a grey background the face which was mostly matched by the participants, and in the ``subtotal'' and ``percentage'' columns the race which was mostly selected. 
For the AA voice, 7 in 10 of the audio samples were matched to WH faces, ranging from 62\% to 81\% of agreement. Only 2 of the audio samples were matched to a AA face 
and 1 audio sample, \emph{AA3}, was matched to an AS face.
However, in the case of the WH voice, all audio samples were matched to a WH face, ranging from 84\% to 98\% of agreement. 
Overall, the results of Study~2 were also unexpected but, at the same time, altogether similar to the results of Study~1 in spite of the difference in the methodology.

\subsection{Study~3: Focus Group with African Americans}

To evaluate the AA voice with African Americans, we run a third, qualitative study, where we also explored two key questions raised by Studies~1 and~2: (1) whether the AA voice had problems which could explain its attribution to a White speaker or face; (2) if not, what could be the reasons for the results of Studies~1 and~2.

\subsubsection{Study~3: Methodology}
\hfill \\
Participants in Study~3 were African American employees of a large IT company recruited from the pool of participants in the focus groups described in section~\ref{sec:ethical_issues}. They were invited to participate in the study by e-mail
and those who agreed were invited to a virtual, 1-hour meeting, asked to sign a standard consent form, and invited to a special, private \emph{Slack} channel. Like before, they were not provided any reward for their participation in Study~3.

The participants in Study~3 went through the following phases:
\begin{description}[labelindent=0cm,leftmargin=0.3cm,style=unboxed]
    \item[Introduction:] Participants were informed about the objectives of the study, the main steps, and asked to briefly introduce themselves..
    \item[Overview of the project:] Participants were given an overview of the project and its current status. 
    \item [Evaluation of the AA voice:] Participants were played 3 audio samples of the AA voice and asked about its quality and resemblance to an African American female voice. They were also shown 3 samples from the actual voice talent and asked for comparisons.
    \item[Exploration of the results of Study~1:] Study~1 was explained to the participants and 
    they were asked to guess the percentage of U.S. English speakers that would label the AA voice as African American. Then they were shown the actual results and asked to comment.
    \item[Exploration of the results of Study~2:] Study~2 was explained to the participants 
    following by an actual evaluation performed in Study~2. They were asked which faces they would consider as a good match for the AA voice and then were shown the actual results and asked to comment. They were also shown two other evaluation samples of the AA voice and one of the WH voice, and asked to comment.
    \item[Exploration of the overall results:] Participants were shown the final overall results of both studies and asked to comment.
    \item[Thanks:] Participants were thanked and asked for suggestions and
comments. 
\end{description}

\subsubsection{Study~3: Results and Findings}
\hfill \\
There were 10 participants in Study~3 (6 female, 4 male), all self-identified African Americans. Each of them is identified as $Pi$ with the number $i$ being assigned randomly. 
The study took 75 minutes, with 15 minutes added by request of the participants. 


When participants were asked to comment on
the 3 audio samples from the AA voice, most of the comments mentioned that the voice was \emph{``pleasant''}, \emph{``natural''}, and \emph{``representative''}. P10 added that \emph{``she kind of sounds like one of my good [African American] friend'' (P10)}, and P9 added that \emph{``I think that you guys kind of hit the mark'' (P9)}. 

When presented with recordings of the actual voice talent for comparison, P10 remarked that \emph{``a little bit higher pitch than the actual voice''} 
while P7 argued that \emph{``[the developers] took a little bit of her Black away. [\ldots   But]  you saved enough of it for me to recognize that's a Black woman speaking'' (P7)}. P9 commented that \emph{``it's more, kind of the inflection of how the original person was talking'' (P9)}, to what P6 replied that \emph{``without a doubt, I knew that [the AA voice] was a Black woman speaking'' (P6)}.


When participants were asked to guess the percentage of U.S. English speakers that they think would label the AA voice as African American in that study, their answers, collected in the \emph{Slack} channel, ranged from 70\% to 90\%, with an average of 79\%. When they were shown the actual result of attribution to the AA voice, which was only 18\% to an African American person, they were quite surprised: 

\begin{quote}
\emph{
``We can recognize  \ldots  Black voices that speak the way this synthetic voice speaks.  \ldots  
So other [non-African American] people who are looking to hear or recognize what they think is an African American voice are looking for people that are splitting verbs and aren't speaking good English'' (P3)}.
\end{quote}

Other participants agreed with P3 in that the results may be caused by the difficulties that non-African Americans have in recognizing African American voices as such. But P3 also raised the issue caused by expectations about the use of Ebonics, and the prejudice which associates African Americans to lack of education, even from their own community.
\begin{quote}
    \emph{
    ``If [the AA voice] was  \ldots  splitting or talking some Ebonics, that didn't sound the words properly or phonetically, then they would probably think that [the AA voice is] like Chris Tucker\footnote{An African American comedian and actor known for its voice impersonations, such as in \url{https://www.youtube.com/watch?v=y_4gaQyqu7o}.} \ldots [This is] the bias, depending on who you were talking to, of what stereotypically an African American is supposed to sound like  \ldots   
    for someone that  \ldots  all they know is TV  \ldots  [the AA voice] probably sounds nothing like [an African American person]'' 
     (P9).} \\
\end{quote}

A discussion followed about whether the goal of the project was to have an African American synthetic voice recognized as so by \emph{all} U.S. English speakers or by the African American community, and the difficulty of meeting the two goals at the same time. P5 asked \emph{``is the idea for me to recognize it, or is it for other people to recognize it?'' (P5)}.
P10 summarized: 
\begin{quote} 
\emph{``I think [the AA voice] works fine, especially if  \ldots  we want to have representation for African Americans  \ldots  We hear ourselves [in the AA voice], but [for] someone from a different race,  \ldots  they can't tell the difference;  \ldots  
all they know is what they've been shown in negative portrayals that are more stereotypical'' (P10).}
\end{quote}

When the participants were shown an example of Study~2, all of them chose AA faces as representative of the AA voice. After being told that the actual results were 93\% for WH faces and only 7\% for AA faces, P10 suggested that, given the results of Study~1, they would expect people to go for 
the face perceived as ``non-White'': \emph{``[the chosen WH face] looks like she might be not necessarily white, but like, maybe from some other countries'' (P10)}. P10 added that \emph{
``we clearly hear an African American, so it's like, how are other people not hearing it?  \ldots  [The participants] are not hearing African American because it's too proper, it's too formal, and it is not how they stereotypically would think [an African American voice] would sound'' (P10)}.

The discussion followed with the participants commenting about their surprise: 
\emph{`` I'm absolutely baffled'' (P10)}; \emph{``That's so crazy  \ldots  they don't know whether it's White or Black, so they default to other [non-White]'' (P7)}. Also, when shown the WH voice, they agreed it sounded White, and P10 added that \emph{``when they are absolutely clear that it's a White voice, they'll they'll choose a White person [a WH face], but when they don't quite know, [they] find the most ambiguous person as a White person'' (P10)}.

After seeing the overall results of studies~1 and~2, P9 suggested that \emph{``I think the question becomes what a Black person is.  [\ldots If a voice] is formal, and it's fully enunciated, and it's clean-sounding, maybe a lot of people don't associate [to an African American voice] because of  \ldots  the stereotypes [of the participants in the study]'' (P9)}. P6 suggested that it could be that \emph{``participants of the study were, predominantly, not African American'' (P6)}. 

At the end of the study, the organizers asked once more whether  \emph{``anyone in this group would have difficulty in recognizing this synthetic voice as African American.''} All participants agreed that the AA voice is distinctively African American, to what P5 added: \emph{``No one that I know would have had an issue recognizing it. No Black person that I know would have had an issue'' (P5)}. 
P10 summarized:
\begin{quote}
    \emph{``The participants [of studies~1 and~2] are just biased to believe that all African Americans speak Ebonics, and are from the hood, and all this stuff.  \ldots  They're not gonna hear [the voice] as African American just because it is not talking that way'' (P10).}
\end{quote}

P1 added 
\emph{``If 
we heard incorrect grammar, that would throw us off. But correct grammar is not what we were listening for. We were listening for texture, intonation, and inflections  \ldots  Maybe some of these other people  \ldots  were listening for something a little more broad than that, a little more stereotypical'' (P1)}. 
P2 then re-emphasized the identity aspects: \emph{``I think what this research study is emphasizing is our cultural references. What am I expecting to hear that qualifies for me as an African American? My African American culture informs me of what is, and what sounds like an African American'' (P2)}.  

At the end, the participants gave the authors some final advice. P9 asked us to not try to increase African American aspects of the voice:
\begin{quote}
    \emph{``When I heard it [the AA voice] sounded authentic  \ldots  
    Please don't feel like you need to put any extra emphasis on, like any colloquialisms in the language, to try to push a Black voice. Don't feel like you have to change how enunciations are happening, or throw some kind of colloquial term in there to try to make it sound Black because she [the AA voice] already sounds Black'' (P9).}
\end{quote}

P10 added:
\begin{quote}
    \emph{``I think it meets the goal  \ldots  if a Black person hears this voice, do they recognize it as Black? Yes. Do we care that anyone else doesn't? No, right, then we're good'' (P10).}
\end{quote}


\subsection{Discussion}


The 3 studies evaluating whether the AA voice sounds as an African American seem to indicate that a generic U.S. audience does not perceive it as such, but African Americans do. The main hypothesis to explain this difference is the one suggested by the participants of Study~3, that most U.S. English speakers have a stereotyped, prejudiced view of what an African American sounds like, including grammatical errors (with respect to SAE), excess of colloquialism, lack of education, and use of Ebonics. Since the AA voice was, by design, a professional-sounding voice trained on a corpus which follows written standard English, and the text fed to it for testing used standard vocabulary and syntax, participants in Studies~1 and~2 
most likely did not detect the linguistic cues which were part of their biased expectations.

Although we were expecting some level of difficulty and confusion in the attribution of the AA voice to an African American person, 
the results of the first two studies were absolute surprising to us. Participants showed a very high level of assertiveness of recognizing AA voice as White, for most audio samples:  74\% White to 15\% African American in the direct question study and 74\% to 20\% in the face matching study.  
However, it was unanimous in Study~3 that the AA voice is distinctively African American, both considering their answers to that question, and the many statements of approval we listed in the findings. Based on that, we are comfortable to assume that, for an audience of African American speakers, the AA voice sounds like an African American person,  and therefore the answer for \textbf{RQ3 is Yes}.

Answering RQ4, whether it is possible to create a TTS system which sounds African American for most U.S. English speakers, is a more complex and multifaceted task, though. As we clearly saw from the results of Studies~1 and~2, a generic U.S. audience 
attributed the AA voice to a White woman while having no difficulty correctly categorizing the WH voice. According to Study~3, the problem was not the AA voice itself but the participants' stereotypical representations of how an African American speaks which is often equated to bad English, colloquialisms, and limited vocabulary.

We concluded that RQ4 can not be answered by using our AA voice given that it seems to defy the (prejudiced) expectations of how an African American sounds by U.S. English speakers at large. Although we believe it is possible, using current technology, to create a synthetic voice with the enunciations and colloquialisms which would match those stereotypes, that would most likely create a serious ethical issue, since we got a clear message from our first focus groups that such a voice would not be welcomed as a representation of an African American person by the African American community. 

So, in our view,  the best answer for 
\textbf{RQ4 is No}, independently of the technology used, because no ethical rendition of an African American voice is likely to pass the prejudiced view of U.S. English speakers about how African Americans speak while satisfying the African American community's own expectations at the same time. Of course, this is likely to change as racial prejudice decreases with time and the differences and similarities among races and ethnicities is better comprehended by people.

However, as suggested by one of the participants in Study~3, maybe we could get more correct attributions of the AA voice in the studies if more information about what constitutes an African American voice was given to the participants, and if they were instructed to focus on intonation, or to consider the voice as coming from an educated person, etc. We are going to explore those issues in future works, but it is possible that even when doing so, the preconceived views about the speech patterns of African Americans will still prevail.

Finally, we believe that the answer to \textbf{RQ5}, whether misconceptions and prejudices from the target audience
affect the evaluation of the ethnicity or race of a synthetic voice, is a cautious \textbf{Possibly}. Our three evaluation studies seem to have shown a case where misconceptions and prejudices from a generic U.S. audience may have affected their evaluation of the race of the AA voice. 
However, we believe we still need to perform more studies to confirm whether prejudice was the key factor here, although neither us nor the participants in the focus group could find other plausible reason for the findings in Studies~1 and~2.

\section{Limitations of the Study}

There are several factors and choices which may impact the findings of this work. First, we performed the initial focus groups and Study~3 with a group of well-educated, middle class, and world-knowledgeable African-Americans, all employees of a large IT company. 
In many ways, they are hardly representatives of the actual African American community in the U.S., and we wonder how the results would be different if we considered a broader spectrum of the population.

Another important limitation is the use of crowd-sourcing platforms for the large voice evaluation studies. As discussed in details in appendix~B, 
we found it very hard to obtain reliable demographic answers from the digital workers, so we have limited confidence in our estimate that the participants of the evaluation Studies~1 and~2 follow the  U.S. demographics. However, that data make us believe that the number of African American participants in the studies is at least not significantly larger than its current demographic share, so we should expect that those participants are, in their vast majority, non-African Americans and therefore likely to have difficulties in recognizing an AA voice.

Finally, we are aware that this work uses many racial and ethnic generalizations and stereotyping which some people may consider inappropriate or even offensive. We tried to be very cautious in the use of terminology and language, but at the same time we tried to avoid excessive use of complex or very technical terms. Even the characterization of a synthetic voice as sounding ``African American'' is problematic given the diversity of voices and the effect of regional patterns of speech. We tried to avoid letting those difficulties lead us to intellectual paralysis and to a lack of action to mitigate the issue of Whiteness of AI, but in the process we may have used some terms in ways which someone might think as incorrect, inappropriate, or offensive. If that was the case, we apologize in advance for them and ask to be contacted so we can correct and improve our terminology.

\section{Final Discussion}

In this paper we described the process of creating an African American-sounding TTS system (referred as the AA voice), starting with the initial study performed to identify ethical issues, determine criteria for voice selection, and select the source talent, followed by the development of the voice, and concluded with 3~studies to evaluate whether the AA voice sounded African American both for a generic U.S. audience and for African Americans.

With the findings from the studies we were able to answer our 5~research questions. We collected 5~key ethical considerations for the development of the AA voice, that is, RQ1: participation of African Americans, diversity of voices, avoidance of stereotyping and Ebonics, difficult of recognition by non-African Americans, and ensuring an appropriate use. The initial study also provided a set of criteria for African American voices, that is, RQ2: representativeness, clarity and speech quality, empathy and expressiveness, and appropriateness for its use. Our evaluation studies showed that the AA voice is recognizable by African Americans as African American, answering positively RQ3. As for the possibility of creating an AA voice which is recognizable by generic U.S. English speakers, our answer to RQ4 is no, since this may require accentuating stereotypical linguistic features and colloquialisms which would reinforce negative stereotypes of African Americans, leading to a violation of the ethical guidelines collected. Finally, we believe misconceptions and prejudice were the main cause of the failure to recognize the AA voice by generic U.S. speakers, but we still need to do further studies, so the answer to RQ5 is a cautious possibly.

In our view, the most interesting contribution of this paper is the exposition of the ethical conflicts involved in addressing inclusiveness in human representations by AI systems. We presented a case where a vocal representation recognized by African Americans was not recognized by non-African Americans, and provided some evidence that a (exaggerated, negative) stereotypical representation more easily recognizable by generic U.S. speakers would not be considered appropriate by African Americans. This was surprising considering that previous literature indicated that recognition in such contexts was easily done~\cite{purnell1999perceptual,scharinger2011you}.

Interestingly, the issue of difficulty of recognition of a non-stereotypical AA voice was identified in the initial focus groups but we were surprised by the level it showed up in the evaluation studies.
We were expecting some level of confusion or non-conclusive results, but the two large-scale studies, with about 570 valid participants, yielded clear results: 62\% of the evaluations considered the sound of the AA voice as White and 58\% matched it to a White face. 

These results can also be understood as posing an ethical dilemma to developers: should they use negative stereotyping to achieve inclusiveness? Or should they follow the positive representations expected by the community and possibly contribute to exclusion? We foresee that a possible solution is to avoid the reliance on single elements to portray a race or ethnic group, as suggested by some comments made during the initial focus groups: \emph{``[Minority kids] don't think that I am Black [when they hear only my voice], and so they are not connected to me, but it's totally different though, when they can see me'' (P3.6)}. Similarly, it is possible to explore stereotyping which are positively viewed by the community, such as the use of deeper tones as suggested by P3.1: \emph{``When I put my James Earl Jones voice  \ldots  then I come off as being African American'' (P3.1)}. We see those ideas as part of our future work, together with a stronger validation and a deeper exploration of whether prejudice was the main cause of the surprising level of failure in recognizing the AA voice as African American.

Finally, we think it is necessary to study the use of the AA voice in the real contexts it was developed for, that is, customer service and tutoring. We are currently exploring methodologies and scenarios to perform such studies considering the use of the voice in contexts where it is both employed with generic U.S. English speakers and within the African American community.

\bibliographystyle{ACM-Reference-Format}
\bibliography{iui24aatts}

\pagebreak

\noindent \textbf{APPENDIXES}


\renewcommand{\thesection}{\Alph{section}} 
\setcounter{section}{0}

\section{Creating a Set of Stereotypical Synthetic Faces}

To perform Study 2, we first had to create 3~sets of disjoint stereotypical AA, WH, and AS female faces. This process started with choosing the \emph{Synthetic Faces High Quality (SFHQ)} dataset~\footnote{https://github.com/SelfishGene/SFHQ-dataset} as the source of images. This dataset comprises more than 425,000 synthetic faces with features associated to race and ethnicity, age, gender, and several other characteristics. A key requirement needed for the success of Study~2 was that the created sets had to be \emph{strongly disjoint}, i.e., that a face belonging to one of the race sets had to be recognized as part of the two other sets with very low likelihood. We discuss some issues and implications related to this requirement later.

To create the 3~strongly disjoint stereotypical AA, WH, and AS female faces we performed the study described next.

\subsection{Methodology}
The goal of this study was to create 3~datasets of 10~synthetic faces each to be used as context for Study~2, where the AA and WH voices would be matched to AA and WH faces and the AS faces were used as distractors.
We first extracted about 200 faces of each race from the SFHQ dataset by searching female faces with an age criterion in the 30 year-old range. We also imposed a constraint regarding facial expression (smiling face).

The authors then went through a manual thinning process where each of them selected about 25~out of the 200~faces from each of the 3~races, considering criteria such as image synthesis quality, presence of smiles (to equalize empathy), and reduced levels of inter-racial ambiguity. For AS faces, we focused on faces from East Asia (China, Japan, North and South Korea, etc.) to avoid possible ambiguities with faces from West Asia (Arab Peninsula, Iran, etc.), North Asia (Russia, Mongolia, etc.) and South Asia (Pakistan, India, Bangladesh, etc.). We also tried to avoid selecting faces which could be recognized as American Indian or Native Hawaiian. For AA and WH faces, we avoided faces which could be also recognized as Hispanic or Latino.

\begin{figure*}[t!]
  \centering
  \includegraphics[width=15cm]{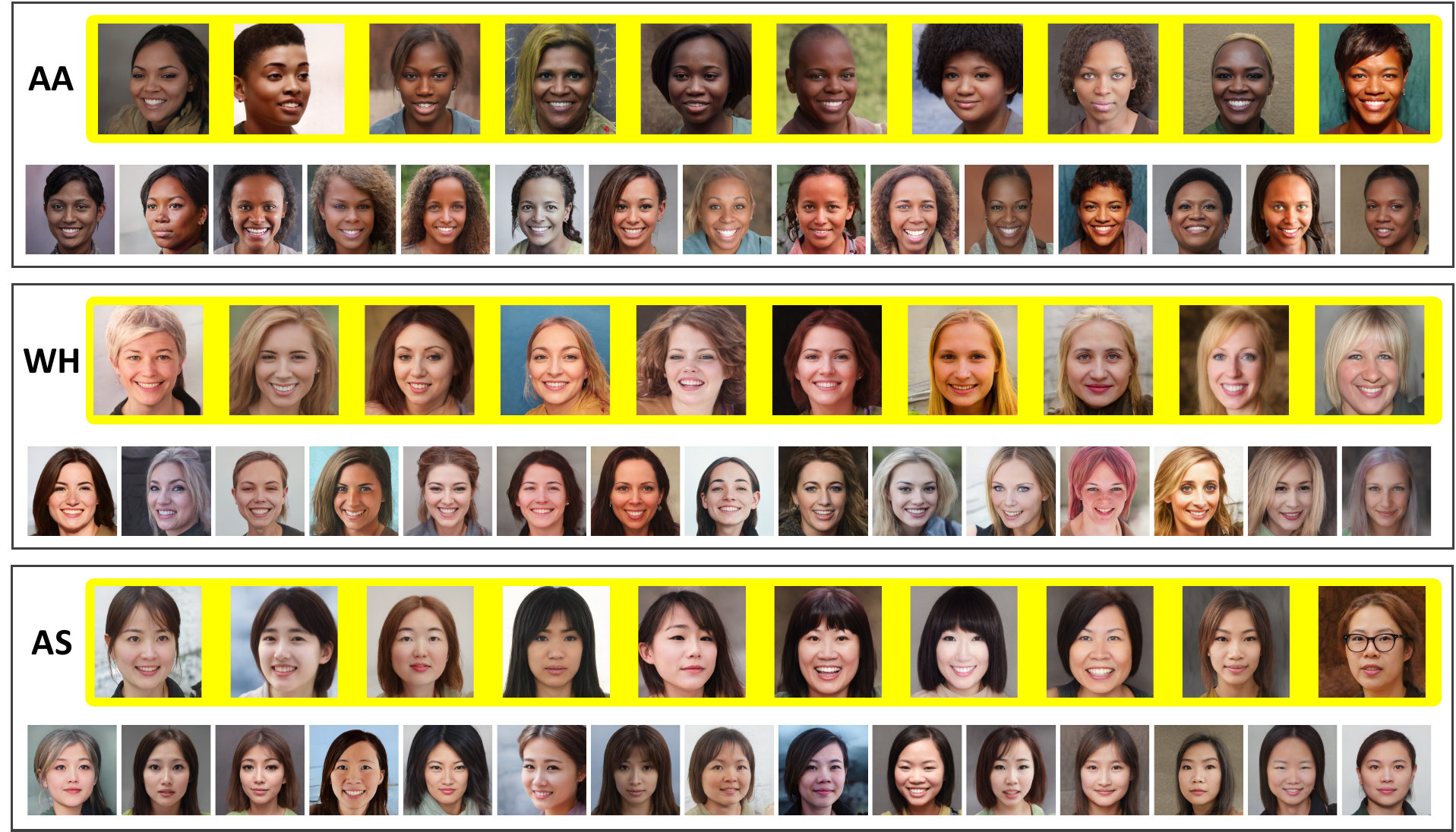}
   \caption{Candidate images for AA, WH, and AS evaluated by the participants in the study; highlighted images were selected to be used in Study~2.}
   \label{fig:app_candidates} 
\end{figure*}

Next, a vote-down scheme was used where all the authors went through the set of faces chosen by all of them and individually removed from consideration faces which any of the authors saw as problematic. We understand that this process could be biased by the authors' own perspectives and prejudices, but we tried to control this issue by pro-actively taking an inclusive and diverse perspective. More importantly, this step aimed to produce only smaller sets of 25 images each, which were then used in a more diverse study with digital workers. Figure~\ref{fig:app_candidates} shows the 3~sets resulting from the thinning process carried out by the digital workers.

Those sets were then submitted to a crowd-sourcing study where participants were asked to attribute a race/ethnicity to individual faces extracted from those sets. 
This attribution was performed by using the race and ethnicity categories defined by the \emph{Standards for the
Classification of Federal Data on Race
and Ethnicity} of the \emph{U.S. Office of Management and Budget (OMB)} in 1997, used in the U.S. Census since 2000\footnote{https://www.govinfo.gov/content/pkg/FR-1997-10-30/pdf/97-28653.pdf}. According to this standard, at least 6 categories for race and ethnicity should be used: \emph{American Indian or Alaska Native}, \emph{Asian}, \emph{Black or African American}, \emph{Hispanic or Latino}, \emph{Native Hawaiian or Other Pacific Islander}, and \emph{White}.

Participants were told that they were evaluating the quality of a synthetic face and, in the process, they were also asked about the race and ethnicity of the face. We decided to disguise the study to minimize the effect of personal opinions about race and ethnicity. 
Participants were equally exposed to faces from the 3~sets.

\begin{figure*}[t!]
  \centering
  \includegraphics[width=15cm]{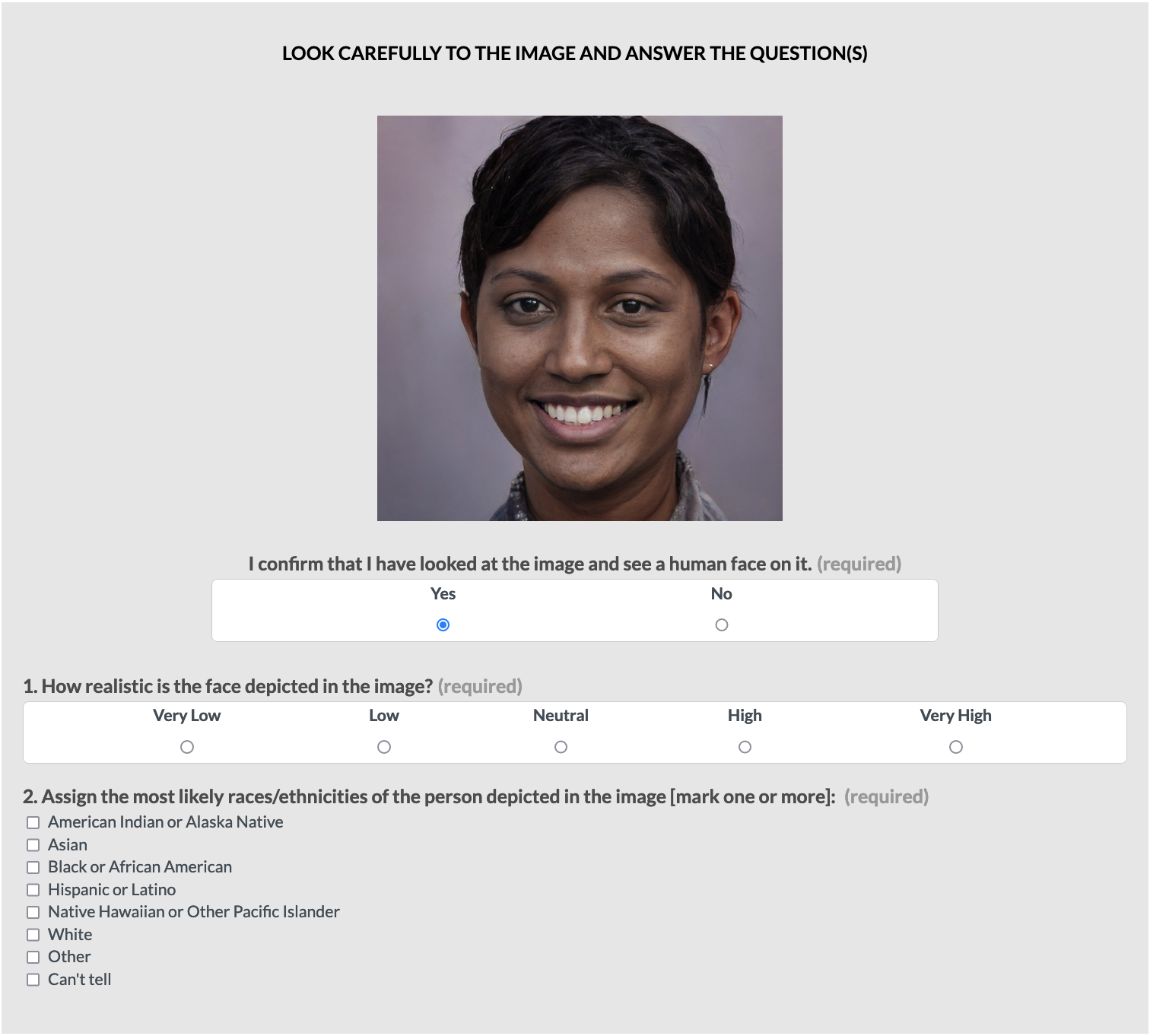}
   \caption{Screen interface of the main task performed by the participants of the face selection study.}
   \label{fig:app_screen_faceselection} 
\end{figure*}

Each participant in this study performed the following sequence of tasks:
\begin{description}
[labelindent=0cm,leftmargin=0.3cm,style=unboxed]
    \item[Introduction:] Participants were informed about the objectives of the study and main steps, provided with an example of the main task, and given rules and tips for success.
    \item[Demographics questionnaire:] Participants were asked to answer demographics questions about gender, age, and race/ethnicity (also using the \emph{Standards for the Classification of Federal Data on Race
    and Ethnicity} of the \emph{U.S. Office of Management and Budget (OMB)}. This task was mandatory, but in each question the participant was allowed to choose a ``prefer not to answer'' option.
    \item[Main task:] Participants performed 6 times a process of seeing a face, evaluating its realism, and choosing its most likely race or ethnicity (with multiple choices allowed). 
    If required responses were missing, the worker was reminded to complete them (see fig.~\ref{fig:app_screen_faceselection}).
    \item[Thanks:] After completing the main task, the workers were thanked and asked for suggestions and comments.
\end{description}

\begin{figure*}[t!]
  \centering
  \includegraphics[width=15cm]{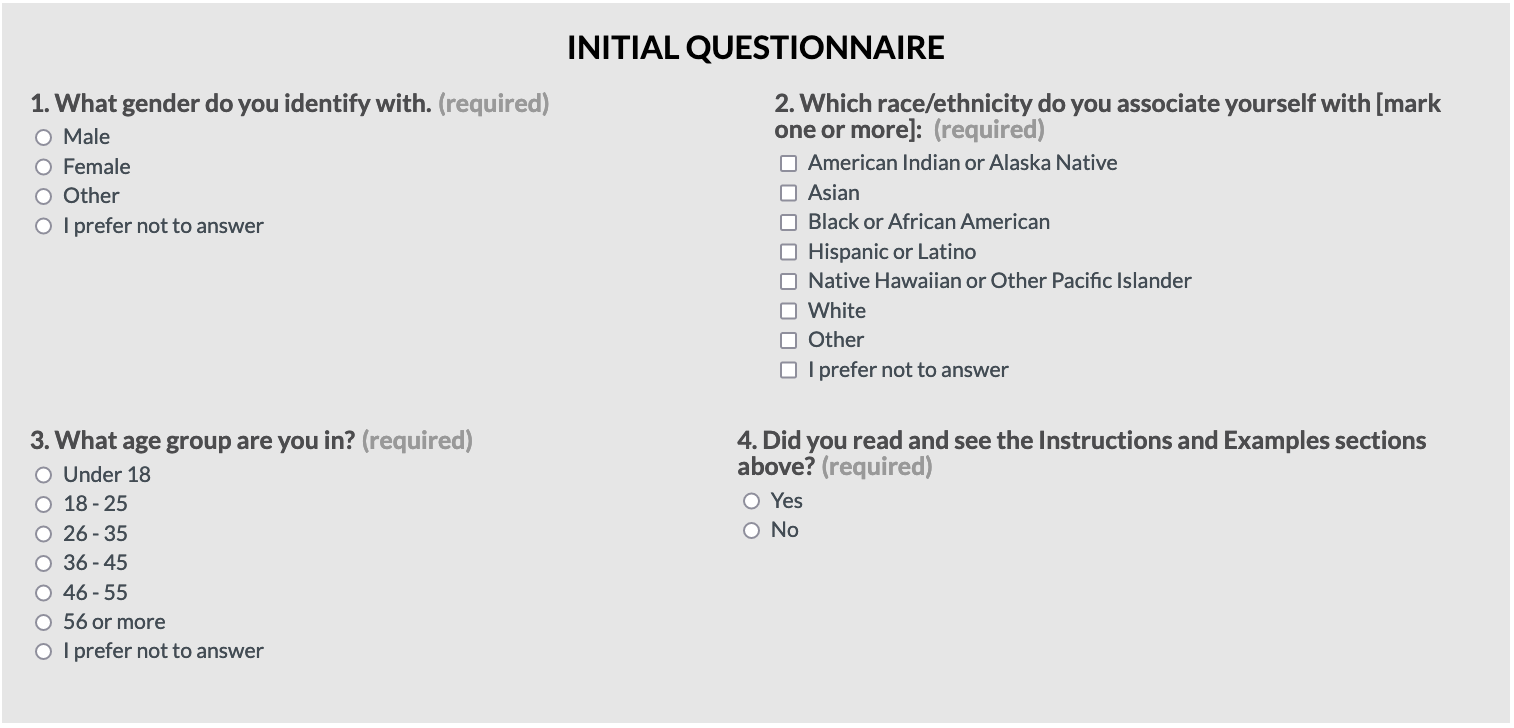}
   \caption{Screen interface of the demographics questionnaire performed by the participants of the face selection study.}
   \label{fig:app_screen_demographics} 
\end{figure*}

Figure~\ref{fig:app_screen_faceselection} shows the screen interface of the main task performed by the participants of this study. Notice that the participant has to first answer the disguise question about the level of realism of the face, and then follow with the multiple-choice question about race and ethnicity. 
Participants were randomly assigned to different faces from the 3~sets (2~from each for each question) and never saw the same face twice.

\begin{table*}[t!]
  \centering
   \caption{Distribution of the racial attribution for each AA, WH, and AS image which was selected.}
   \label{tab:app_statistics_selected}   \includegraphics[width=13.5cm]{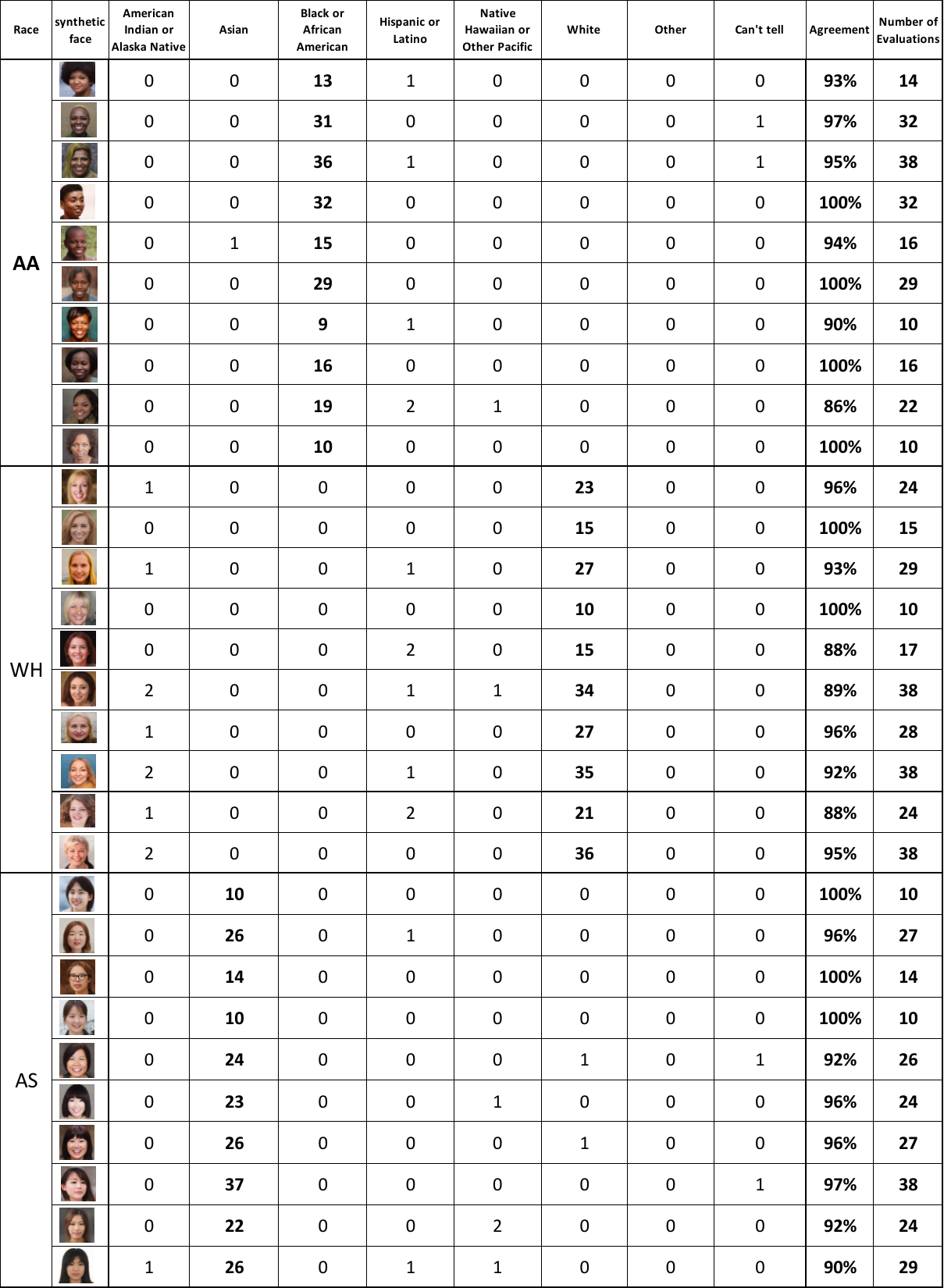}

\end{table*}

\begin{table*}[t!]
  \centering
  \caption{Distribution of the self-reported race/ethnicity of the participants of the image selection study described in appendix~A compared to the U.S. 2020 census adult data.}
  \label{tab:selected_image_statistics} 
  \includegraphics[width=12cm]{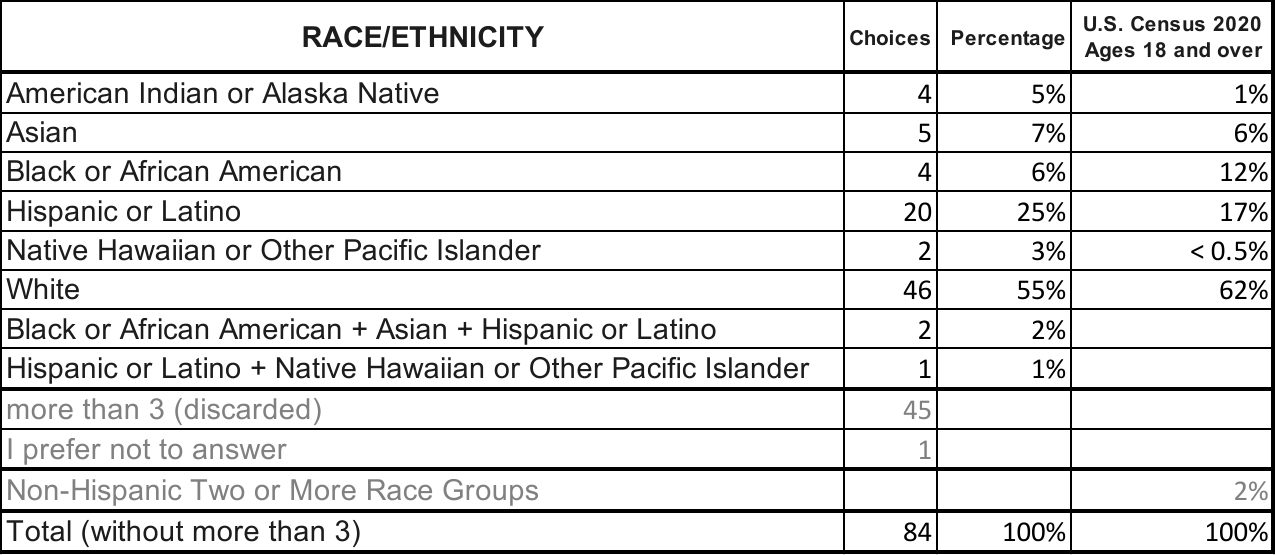}
  \setlength{\belowcaptionskip}{5pt}

\end{table*}

\subsection{Results}

There were 130 participants (79 female, 48 male, 3 other) in the study, which provided 1092 evaluations of the 75 faces. The demographics of the participants is shown in detail in table~\ref{tab:selected_image_statistics} and discussed later. Given a candidate face, we established as criterion for its acceptance that the corresponding race be solely attributed by at least 9 out of 10 participants, or by more than 85\% if there were more than~15 evaluations. As faces reached that level of attribution or above, they were removed from the pool of faces being evaluated, and the study was kept running until at least 10~faces from each race met the criterion. In average, the faces selected had an agreement level of 95\% in more than 700~evaluations. This led to the set of 10 faces highlighted (in yellow) for AA, WH, and AS in fig.~\ref{fig:app_candidates}.

Table~\ref{tab:app_statistics_selected} depicts the distribution of the racial attribution for the 10~selected faces of each race, following the criterion defined before. As mentioned, this table shows only the faces in which we achieved a high level of agreement of attribution. This was not the case for many of the faces depicted in fig.~\ref{fig:app_candidates}. For instance, the leftmost AA face of the non-selected row (that is, not highlighted in yellow) of fig.~\ref{fig:app_candidates} was a case where 10~participants considered it as Black or African American, other 10~as Hispanic or Latino, and 6~selected other options. Similarly, the 7\textsuperscript{th}~non-selected WH face had 16~participants selecting it as Hispanic or Latino, 12~as White, and 1~as Native Hawaiian or Other Pacific Islander.

\subsection{Discussion}

It is important to notice that the goal of this study was not to create a set of faces for each race which was representative of the facial diversity of the particular group. Rather, the purpose was to find a set of faces which were easily recognizable as belonging to that race and not recognizable as belonging to any of other racial/ethical groups according to the categories used in the U.S. Census and, particularly, to the other two sets. In that sense, each set is not only a set of stereotypical faces of that race but a set which is perceived as disjoint to the other sets.

We are plainly aware of how problematic the results of this study can be if we were considering the general case or the use of the constructed sets in other contexts. We do not recommend the sets we created to be used in other studies and contexts, since the sets were built with a specific goal: to provide stereotypical, racially-disjoint sets of images of faces to be matched to synthetic voices in the context of the voice evaluation Study~2. Therefore, stereotyping and avoidance of characteristics from multiple races/ethnicities were desired features of the sets given the purpose of using the sets in the voice attribution study. Therefore, other uses of these sets of faces are disallowed by us.

\section{Demographics of the Participants of Studies 1 and 2}

Table~\ref{tab:selected_image_statistics} shows the racial/ethnic distribution of the participants in the image selection study described in appendix~A, as self-reported in the demographics questionnaire task (see fig.~\ref{fig:app_screen_demographics}). Of the 130~participants, only one preferred not to answer and 45~provided more than 3~answers. We discarded both cases from the demographics analysis (about 33\% of the participants) since it is quite possible that participants who indicated more than 3 options were, in fact, not willing to disclose their race/ethnicity. Even if this was not the case, the contribution of them to the analysis of the distribution of the demographics would be dispersed.

Table~\ref{tab:selected_image_statistics} also lists the percentages of the different races/ethnicities according to the U.S. census of 2020\footnote{As reported in \url{https://datacenter.aecf.org/data/bar/6539-adult-population-by-race-and-ethnicity}.}. In this study, the percentage of Black or African American is a little lower than the U.S. demographics, 6\% to 12\%, respectively; White is slightly lower, 55\% to 62\%; and a little higher for Hispanic or Latino, 25\% to 17\%. Those variations are reasonable, in our view, given the size of the filtered sample (84 participants), the fact that we had to discard about one third of the answers, and the difficulties people have self-reporting race/ethnicity~\footnote{As discussed, for instance, in~\url{https://www.pnas.org/doi/abs/10.1073/pnas.2117940119}.}.

Considering the results and the discussion above, we consider it is reasonably safe to assume, for the purposes of this study and other similar studies conducted in the same \emph{Appen} platform, as Studies~1 and~2 described in the paper, that the demographics of the participants is similar to the U.S. adult demographics.

\end{document}